\documentclass[runningheads]{llncs}

\usepackage{hyperref}
\usepackage{url}

\usepackage[utf8]{inputenc} 
\usepackage[T1]{fontenc}    
\usepackage{hyperref}       
\usepackage{url}            
\usepackage{booktabs}       
\usepackage{amsfonts}       
\usepackage{nicefrac}       
\usepackage{microtype}      
\usepackage{amsmath}
\usepackage{bm}
\usepackage{eurosym}

\usepackage{mathtools}
\DeclarePairedDelimiter\abs{\lvert}{\rvert}%

\makeatletter
\let\oldabs\abs
\def\abs{\@ifstar{\oldabs}{\oldabs*}}
\let\oldnorm\norm
\def\norm{\@ifstar{\oldnorm}{\oldnorm*}}
\makeatother

\usepackage{subfig}

\usepackage[pdftex]{graphicx}
\usepackage[misc]{ifsym}

\spnewtheorem{definitiontheorem}{Definition}[section]{\bfseries}{\itshape}

\begin{document}

\title{SoccerMap: A Deep Learning Architecture for Visually-Interpretable Analysis in Soccer}
\titlerunning{SoccerMap: An Architecture for Visually-Interpretable Analysis in Soccer.}

\toctitle{SoccerMap: A Deep Learning Architecture for Visually-Interpretable Analysis in Soccer}

\newcommand{\fix}{\marginpar{FIX}}
\newcommand{\new}{\marginpar{NEW}}

\tocauthor{Javier Fern\'{a}ndez (Polytechnic University of Catalonia and FC Barcelona), Luke Bornn (Simon Fraser University)
}

\author{Javier Fern\'{a}ndez\inst{1,2}\Letter\and
Luke Bornn\inst{3}}
\authorrunning{J. Fern\'{a}ndez and L. Bornn}

\institute{Polytechnic University of Catalonia, Barcelona, Spain, \email{javier.fernandez.de.la.rosa@upc.edu} \and
FC Barcelona, Barcelona, Spain, \email{javier.fernandezr@fcbarcelona.cat} \and
Simon Fraser University, British Columbia, Canada,
\email{lbornn@sfu.ca}}


\maketitle

\begin{abstract}

We present a fully convolutional neural network architecture that is capable of estimating full probability surfaces of potential passes in soccer, derived from high-frequency spatiotemporal data. The network receives layers of low-level inputs and learns a feature hierarchy that produces predictions at different sampling levels, capturing both coarse and fine spatial details. By merging these predictions, we can produce visually-rich probability surfaces for any game situation that allows coaches to develop a fine-grained analysis of players' positioning and decision-making, an as-yet little-explored area in sports. We show the network can perform remarkably well in the estimation of pass success probability, and present how it can be adapted easily to approach two other challenging problems: the estimation of pass-selection likelihood and the prediction of the expected value of a pass. Our approach provides a novel solution for learning a full prediction surface when there is only a single-pixel correspondence between ground-truth outcomes and the predicted probability map. The flexibility of this architecture allows its adaptation to a great variety of practical problems in soccer. We also present a set of practical applications, including the evaluation of passing risk at a player level, the identification of the best potential passing options, and the differentiation of passing tendencies between teams.

\end{abstract}

\section{Introduction}

Sports analytics is a fast-growing research field with a strong focus on data-driven performance analysis of professional athletes and teams. Soccer, and many other team-sports, have recently benefited from the availability of high-frequency tracking data of both player and ball locations, facilitating the development of fine-grained spatiotemporal performance metrics \cite{rein2016big}.
One of the main goals of performance analysis is to answer specific questions from soccer coaches, but to do so we require models to be robust enough to capture the nuances of a complex sport, and be highly interpretable so findings can be communicated effectively. In other words, we need models to be both accurate and also translatable to soccer coaches in visual terms.\\

The majority of existing research in soccer analytics has focused on analyzing the impact of either on-ball events, such as goals, shots, and passes,
or the effects of players' movements and match dynamics \cite{gudmundsson2017spatio}. Most modeling approaches share one or more common issues, such as: heavy use of handcrafted features, little visual interpretability, and coarse representations that ignore meaningful spatial relationships. We still lack a comprehensive approach that can learn from lower-level input, 
exploit spatial relationships on any location, and provide accurate predictions of observed and unobserved events at any location on the field.\\

The main contributions of our work are the following: 

\begin{itemize}
\item We present a novel application of deep convolutional neural networks that allows calculating full probability surfaces for developing fine-grained analysis of game situations in soccer. This approach offers a new way of providing  coaches with rich information in a visual format that might be easier to be presented to players than the usual numerical statistics.
\item We show how this architecture can ingest a flexible structure of layers of spatiotemporal data, and how it can be easily adapted to provide practical solutions for challenging problems such as the estimation of pass probability, pass selection likelihood and pass expected value surfaces.
\item We present three novel practical applications derived from pass probability surfaces, such as the identification of optimal passing locations, the prediction of optimal positioning for improving pass probability, and the prediction of team-level passing tendencies.
\end{itemize}

The presented approach successfully addresses the challenging problem of estimating full probability surfaces from single-location labels, which corresponds to an extreme case of weakly-supervised learning.





\section{Related Work} 

From an applied standpoint, our work is related to several other approaches aimed at estimating pass probabilities and other performance metrics derived from spatiotemporal data in soccer. Regarding the technical approach, we leverage recent findings on weakly-supervised learning problems and the application of fully convolutional neural networks for image segmentation.

\paragraph{Soccer analytics} Pass probability estimation has been approached in several ways. A physics-based model of the time it takes each player to reach and control the ball has been used to derive pass probabilities on top of tracking data \cite{spearman2017physics}. Other approaches include the use of dominant regions to determine which player is most likely to control the ball after a pass \cite{gudmundsson2017spatio} or using a carefully selected set of handcrafted features to build linear prediction models \cite{power2017not}. The related problem of pass selection has been approached by applying  
convolutional neural networks that predict the likelihood of passing to a specific player on the attacking team\cite{hubavcek2018deep}. The estimation of pass value has been approached either by the expert-guided development of algorithmic rules \cite{cakmak2018computational}, the application of standard machine learning algorithms on a set of handcrafted features \cite{power2017not}, 
or problem-specific deep learning models with dense layers and single output prediction \cite{fernandezdecomposing}. While some of the related work has estimated probability surfaces by inference on a set of discrete pass destination locations \cite{spearman2017physics,fernandezdecomposing}, none has 
yet approached the learning of probability surfaces directly.

\paragraph{Fully convolutional networks and weakly-supervised learning} Fully convolutional networks have been extensively applied to semantic image segmentation, specifically for the pixel-labeling problem  to successfully detect broad pixel areas associated with objects in images. The approach most related to our work builds a hierarchy of features at different sampling levels that are merged to provide segmentation regions that preserve both fine and coarse details \cite{long2015fully}. From a learning perspective, image segmentation has been approached as either supervised \cite{long2015fully}, weakly-supervised \cite{pathak2015constrained}, and semi-supervised learning problems \cite{papandreou2015weakly}. 
Commonly, available labels are associated with many other pixels in the original image. However, in our case, labels are only associated with a single location in the desired probability map, transforming our learning problem into an unusual case of weakly-supervised learning.

\section{A Deep Model for Interpretable Analysis in Soccer}

\begin{figure}[h!]
  \centering  
  \includegraphics[width=0.99\linewidth]{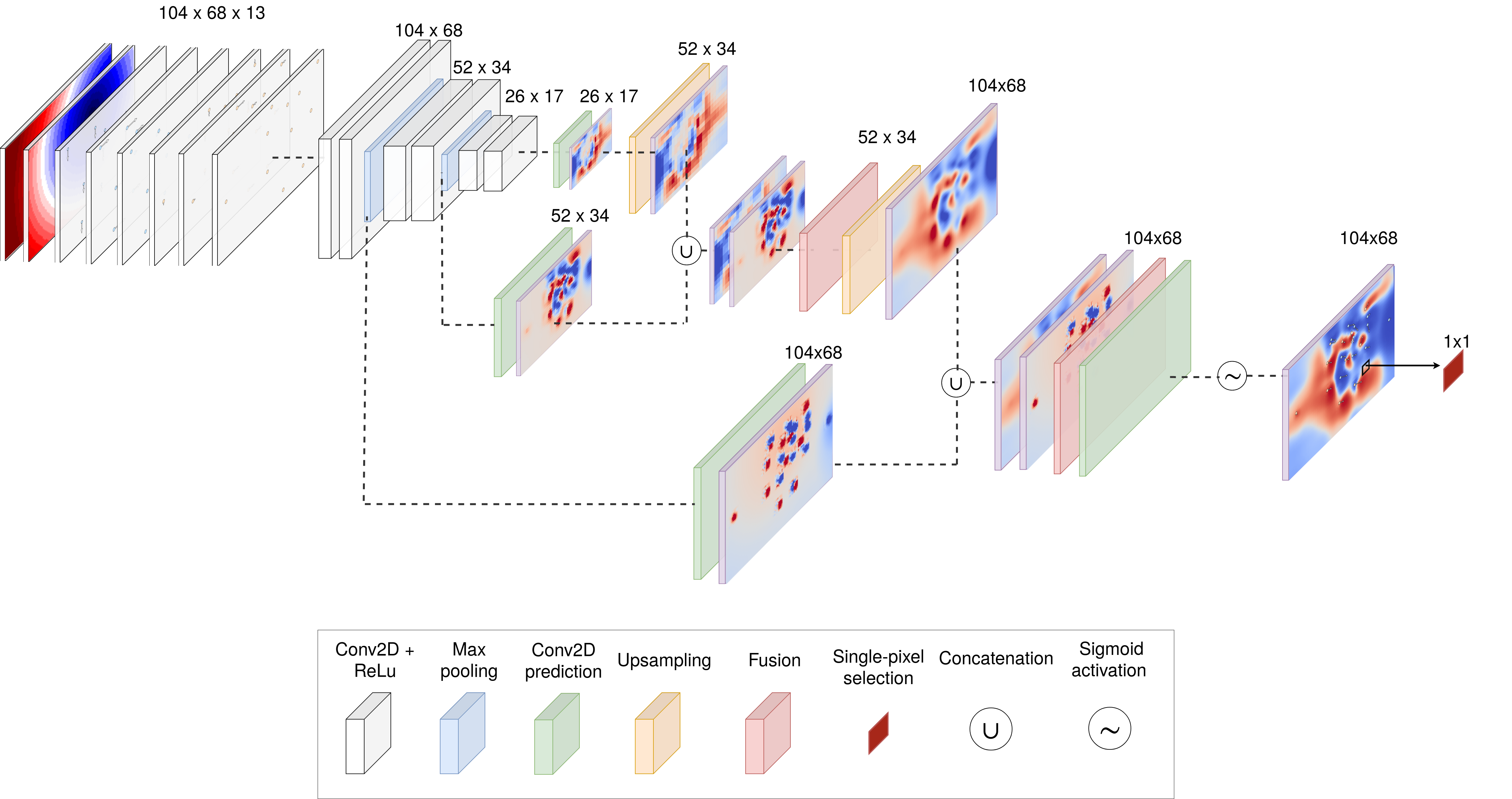}
  \caption{SoccerMap architecture for a coarse soccer field representation of $104\times 68$ and 13 input channels.}
  \label{fig:architecture}
\end{figure}

We build our architecture on top of tracking data extracted from videos of professional soccer matches, consisting of the 2D-location of players and the ball at 10 frames per second, along with manually tagged passes. At every frame we take a snapshot of the tracking data and create 
a representation of a game situation consisting of a $l \times h \times c$ matrix, where $c$ channels of low-level information are mapped to a $l \times h$ coarse spatial representation of the locations on the field. 
 We seek an architecture that can learn both finer features at locations close to a possible passing destination and features considering information on a greater spatial scale. For passes, local features might be associated with the likelihood of nearby team-mates and opponents reaching the destination location and information about local spatial pressure. On the other hand, higher scale features might consider player's density and interceptability of the ball in its path from the location of origin. Finally, we seek to estimate this passing probability to any other location on the $l \times h$ spatial extent of the field.\\
 
This game state representation is processed by the deep neural network architecture presented in Figure \ref{fig:architecture}. The network creates a feature hierarchy by learning convolutions at $1x$, $1/2x$, and $1/4x$ scales while preserving the receptive field of the filters. Predictions are produced at each of these scales, and then upsampled nonlinearly and merged through fusion layers. A sigmoid activation layer is applied to the latest prediction to produce pass probability estimations at every location, preserving the original input scale. During training, a single-location prediction, 
associated with the destination of a sample pass is selected to compute the log-loss that is backpropagated to adjust the network weights. 


\subsection{The Reasoning Behind the Choice of Layers}

The network incorporates different types of layers: max-pooling, linear, rectified linear unit (ReLu) and sigmoid activation layers, and 2D-convolutional filters (conv2d) for feature extraction, prediction, upsampling and fusion. In this section we present a detailed explanation of the reasoning behind the choice of layers and the design of the architecture.

\paragraph{Convolutions for feature extraction}
At each of the $1x$, $1/2x$, and $1/4x$ scales two layers of conv2d filters with a $5\times 5$ receptive field and stride of $1$ are applied, each one followed by a ReLu activation function layer, in order to extract spatial features at every scale. In order to keep the same 
dimensions after the convolutions we apply symmetric padding to the input matrix of the convolutional layer. We chose symmetric-padding to avoid border-image artifacts that can hinder the predicting ability and visual representation of the model.

\paragraph{Fully convolutional network}

There are several conceptual and practical reasons for considering  convolutional neural networks (convnets) for this problem. Convolutional filters are designed to recognize the relationships between nearby pixels, producing features that are spatially aware. Convnets have been proven successful in data sources with a Euclidean structure, 
such as images and videos, so a 2D-mapping of soccer field location-based information can be expected to be an ideal data structure for learning essential features. Also, these features are expected to be non-trivial and complex. Convnets have been proven to learn what are sometimes more powerful visual features than handcrafted ones, even given large receptive fields and weak label training \cite{long2014convnets}. Regarding the architecture design, we are interested in learning the full $l \times h$ mapping of 
passing probabilities covering the extent of a soccer field, for which fully convolutional layers are more appropriate than classical neural networks built for classification when changing dense prediction layers for 1x1 convolution layers.

\paragraph{Pooling and upsampling}





The network applies downsampling twice through max-pooling layers to obtain the $1/2x$ and $1/4x$ representations. Since activation field size is kept constant after every downsampling step, the network can learn filters of a wider spatial extent, leading to the detection of coarse details.
We learn non-linear upsampling functions at every upsampling step by first applying a 
$2x$ nearest neighbor upsampling and then two layers of convolutional filters. The first convolutional layer consists of $32$ filters with a $3 \times 3$ activation field and stride $1$, followed by a ReLu activation layer. The second layer consists of $1$ layer with a  $3 \times 3$ activation field and stride $1$, followed by a linear activation layer. This upsampling strategy has been shown to provide smoother outputs and to avoid artifacts that can be usually found in the application transposed convolutions \cite{odena2016deconvolution}. 

\paragraph{Prediction and fusion layers}

Prediction layers consist of a stack of two convolutional layers, the first with $32$ $1\times 1$ convolutional filters followed by 
an ReLu activation layer, and the second consists of one $1\times 1$ convolutional filter followed by a linear activation layer. 
Instead of reducing the output to a single prediction value, we keep the spatial dimensions at each step and 
use $1\times 1$ convolutions to produce predictions at each location. The stack learns a non-linear prediction on top of the output of convolutional layers. 
To merge the outputs at different scales, we concatenate the pair of matrices and pass them through a convolutional layer of one $1\times 1$ filter.

\subsection{Learning from Single-Location Labels}

We seek a model that can produce accurate predictions of the pass probability to every location on a $l \times h$ coarsed representation of a soccer field, given a $l \times h \times c$ representation of the game situation at the time a pass is made. 
In training, we only count with the manually labeled location of the pass destination and a binary label of the outcome of the pass. 

\begin{definition}[SoccerMap] 
\label{theorem:SoccerMap}
Let $X = \{x | x \in \mathbb{R}^{l \times h \times c}\}$ be the set of possible game state representations at any given time, where $l,h \in \mathbb{N}_1$ are the height and length of a coarse representation of  soccer field,  and $c \in {\mathbb{N}_1}$ the number of data channels, a SoccerMap is a function  $f(x;\theta), f: \mathbb{R}^{l\times h \times c} \to \mathbb{R}^{l \times h}_{[0,1]}$, where $f$ produces a pass probability map, and $\theta$ are the network parameters.
\end{definition}





\begin{definition}[Target-Location Loss]
\label{def:target_location_loss}

Given the sigmoid function $\sigma(x) = \frac{e^x}{e^x+1}$ and let $y_k \in \{0,1\}$ be the outcome of a pass at time $t(x_k)$, for a game state  $x_k$, $d_k$ the destination location of the pass $k$, $f$ a SoccerMap  with parameters $\theta$, and $logloss$ the log-loss function, we define the target-location loss as 
$$L(f(x_k;\theta),y_k,d_k)= logloss(f(x_k;\theta)_{d_k}, y_k)$$
\end{definition}

We approach the training of the model as a weakly-supervised learning task, where the ground truth labels only correspond to a single location in the full mapping matrix that needs to be learned. 
The target-location loss presented in Definition \ref{def:target_location_loss} essentially shrinks the output of a SoccerMap  $f$ to a single prediction value by selecting the prediction value at the destination of the pass, and then computes the log-loss between this single prediction and the ground-truth outcome value. 

\subsection{Spatial and Contextual Channels from Tracking Data}
\label{sec:channels}



Our architecture is designed to be built on top of two familiar sources of data for sports analytics: tracking data and event data. Tracking data consists of the location of the players and the ball at a high frequency-rate. Event-data corresponds to manually labeled observed events, such as passes, shots, and goals, including the location and time of each event. 
We normalize the players' location and the ball to ensure the team in possession of the ball attacks from left to right, thus standardizing the representation of the game situation. On top of players' and ball locations, we derive low-level input channels, including spatial (location and velocity) and contextual information (ball and goal distance). Channels are represented by matrices of $(104,68)$ where each cell approximately represents $1m^2$ in a typical soccer field.

\begin{definition}[Tracking-Data Snapshot]
Let $Z_p(t),Z_d(t),Z_b(t), Z_g(t) \in \{z | z \in \mathbb{R}^{l \times h}\}$ be the locations of the attacking team players, the location of the defending team players, the location of the ball, and the location of the opponent goal, respectively, at time $t$, then a tracking-data snapshot at time $t$ is defined as the 4-tuple $Z(t)=(Z_p(t),Z_d(t),Z_b(t),Z_g(t))$.
\end{definition}

In order to create a game state representation $X(t) \in X$ as described in Definition \ref{theorem:SoccerMap} we produce 13 different channels on top of each tracking-data snapshot $Z$ where a pass has been observed, which constitute the game-state representation for the pass probability model.
Each channel corresponds to either a sparse or dense matrix of size $(h,l)$, according to the chosen dimensions for the coarse field representation. The game-state representation is composed of the following channels: 
\begin{itemize}
\item Six sparse matrices with the location, and the two components of the velocity vector for the players in both the attacking team and the defending team, respectively.

\item Two dense matrices where every location contains the distance to the ball and goal location.

\item Two dense matrices containing the sine and cosine of the angle between every location to the goal and the ball location, and one dense matrix containing the angle in radians to the goal location.

\item Two sparse matrices containing the sine and cosine of the angle between the velocity vector of the ball carrier and each of the teammates in the attacking team.
\end{itemize}

\section{Experiments and Results}

\subsection{Dataset}

We use tracking-data, and event-data from 740 English Premier League matches from the 2013/2014 and 2014/2015 season, provided by \emph{STATS LLC}. Each match contains the $(x,y)$ location for every player, and the ball sampled at $10$Hz. The event-data provides the location, time, player, team and outcome for 433,295 passes. From this data, we extract the channels described in Section \ref{sec:channels} for a coarse $(104,68)$ representation of a soccer field to obtain a dataset of size $433295 \times 104 \times 68 \times 13
$. There are 344,957 successful passes and 88,338 missed passes.

\subsection{Benchmark Models}
\label{sec:benchmark}

We compare our results against a series of benchmark models of increasing levels of complexity. We define a baseline model \emph{Naive} that for every pass outputs the known average pass completion in the full dataset ($80\%$) following a similar definition in \cite{power2017not}.
We build two additional models \emph{Logistic Net} and \emph{Dense2 Net} based on a set of handcrafted features built on top of tracking-data.  Logistic Net is a network with a single sigmoid unit, and {Dense2 Net is a neural network with two dense layers followed by ReLu activations and a sigmoid output unit.

\paragraph{Handcrafted features} We build a set of spatial features on top of tracking-data based on location and motion information on players and the ball that is similar to most of the features calculated in previous work on pass probability estimation \cite{power2017not,spearman2017physics,gudmundsson2017spatio}. We define the following set of handcrafted features from each pass: origin and destination location, pass distance, attacking and defending team influence at both origin and destination, angle to goal at origin and destination, and the maximum value of opponent influence in a straight line between origin and destination. The team's spatial influence values are calculated following the model presented in \cite{fernandez2018wide}.

\subsection{Experimental Framework}
In this section, we describe the experimental framework for testing the performance of the proposed architecture for the pass success probability estimation problem.

\paragraph{Training, validation, and test set} We randomly selected matches from both available seasons and split them into a training, validation, and test set with a $60:20:20$ distribution. We applied a stratified split, so the successful/missed pass class ratio remains the same across datasets. The validation set is used for model selection during a grid-search process. 
The test set is left as hold-out data, and results are reported on performance for this dataset. For the benchmark models, datasets are built by extracting the features described in Section \ref{sec:benchmark}, and an identical split is performed. Features are standardized column-wise by subtracting the mean value and dividing by the standard deviation.


\paragraph{Optimization} 
Both the SoccerMap network and the baseline models are trained using adaptive moment estimation (Adam). Model selection is achieved through grid-search on learning rates of $10^{-3}$, $10^{-4}$ and $10^{-5}$, and batch sizes of $1$, $16$ and $32$, while $\beta_1,\beta_2$ are set to $0.9$ and $0.999$, respectively. We use early stopping with a minimum delta rate of $0.001$. Optimization is computed on a single Tesla M60 GPU and using Tensorflow 1.5.0. During the optimization, the negative log-loss is minimized.

\paragraph{Metrics}
Let $N$ be the number of examples in the dataset, $y$ the ground-truth labels for pass events and $\hat{y}$ the model predictions. We report the negative log-loss $$\mathcal{L}(\hat{y},y) = - \frac{1}{N} \sum_i y_i \cdot log(\hat{y_i}) + (1-y_i) \cdot log(1-\hat{y_i}).$$ In order to validate the model calibration we use a variation of the expected calibration error (ECE) presented in \cite{guo2017calibration} which computes the expected difference between accuracy and confidence of the model on a finite set of samples split into $K$ bins of size $1/K$, according to the predicted confidence or probability for every sample. 
Since our model is not designed for classification, we use the count of the number of examples of the positive class rather than accuracy for $ECE$.

Let $B_k$ be a bin where $k \in [1,K]$ then $$ECE = \sum_{k=1}^{K} \frac{\abs{B_k}}{N} \abs{ \bigg(\frac{1}{|B_k|} \sum_{i \in B_k}1(y_i=1)\bigg) - 
\bigg(\frac{1}{|B_k|} \sum_{i \in B_k} \hat{y}_i  \bigg)  }. $$

A perfectly calibrated model will have a ECE value of $0$. Additionally, we provide a calibration reliability plot \cite{guo2017calibration} showing the mean confidence for every bin $B_k$.

\subsection{Results}

Table \ref{table:results} presents the results for the benchmark models and SoccerMap for the pass probability dataset. We can observe that SoccerMap achieves remarkably lower error than the other models and produces a calibrated estimation of pass probability. Despite the considerably large number 
of parameters in SoccerMap, the inference time for a single sample is low enough to produce a real-time estimation for frame rates below 200Hz. Figure \ref{fig:calibration_plot} presents a calibration reliability plot for each of the models. Both Logistic Net and SoccerMap produce well-calibrated estimations of pass probabilities, however, SoccerMap is proven to be considerably more precise as shown by the difference in log-loss between both.

\begin{table}[h!]
  \caption{Results for the benchmark models and SoccerMap for the pass probability dataset.}
  \label{table:results}
  \centering
  \begin{tabular}{|c|c|c|c|c|}
    \hline
    Model     & Log-loss     & ECE &  Inference time & Number of parameters\\
	\hline
    Naive & $0.5451$ & $-$ & $-$ & $0$\\
    Logistic Net & $0.384$ & \bm{$0.0210$} & $0.00199$s  & $11$\\
    Dense2 Net & $0.349$ & $0.0640$ & $0.00231$s  & $231$\\
    SoccerMap & \bm{$0.217$} & \bm{$0.0225$} & $0.00457$s  & $401,259$\\
	\hline  
\end{tabular}
\end{table}

\begin{figure}[h!]
\centering
\includegraphics[width=0.90\linewidth]{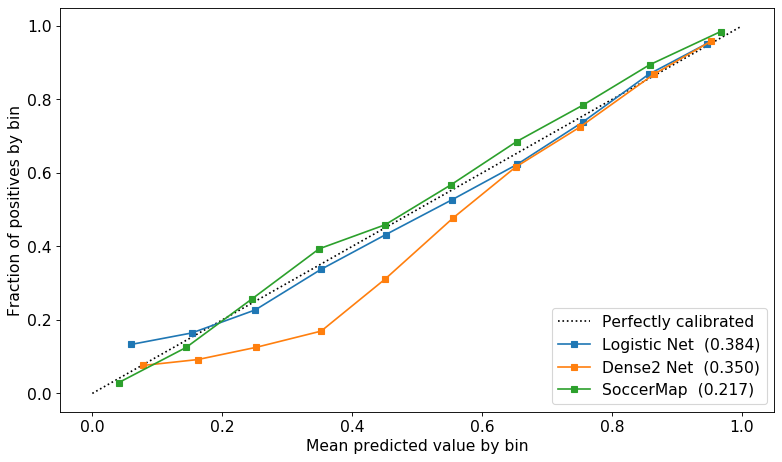}
\caption{A calibration reliability plot, where the X-axis presents the mean predicted value for samples in each of 10 bins, and the Y-axis the fraction of samples in each bin containing positive examples.}
\label{fig:calibration_plot}
\end{figure}

\begin{figure}[h!]
\centering
\includegraphics[width=0.90\linewidth]{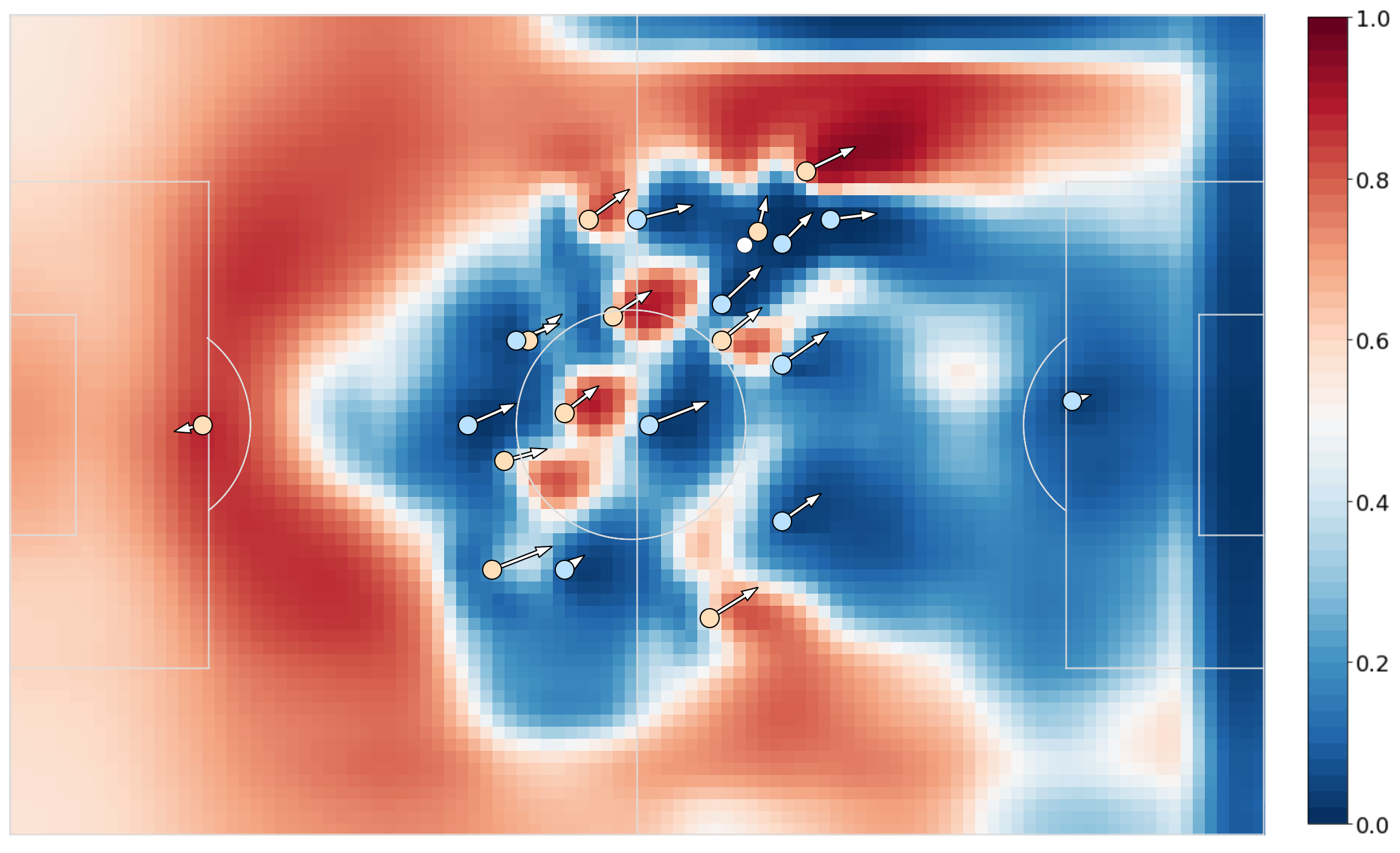}
\caption{Pass probability surface for a given game situation. Yellow and blue circles represent players' locations on the attacking and defending team, respectively, and the arrows represent the velocity vector for each player. The white circle represents the ball location.}
\label{fig:video_and_2d_surfaces}
\end{figure}

Figure \ref{fig:video_and_2d_surfaces} presents the predicted pass probability surface for a specific game situation during a professional soccer match. We observe that the model can capture both fine-grained information, such as the influence of defending and attacking players on nearby locations and coarse information such as the probability of reaching more extensive spatial areas depending on the distance to the ball and the proximity of players. We can also observe that the model considers the player's speed for predicting probabilities of passing to not-yet occupied spaces, a critical aspect of practical soccer analysis.

\subsubsection{Ablation Study}
We performed an ablation study  in order to evaluate whether the different components of the proposed architecture allow improving its performance on the pass probability estimation problem or not, by testing the performance of different variations of the architecture.

Table \ref{table:ablation} presents the log-loss obtained on different configurations of the architecture with the following components: skip-connections (SC), non-linear up-sampling (UP), fusion layer (FL), non-linear prediction layer (NLP), and the number of layers of convolutional filters by sampling layer (NF). We can observe there are two configurations with similar log-loss: the SoccerMap and SoccerMap-UP configurations. While the removal of the non-linear upsampling slightly increases the performance, it produces visual artifacts that are less eye-pleasing when inspecting the surfaces. Given that the surfaces are intended to be used by soccer coaches in practice, SoccerMap provides a better option for practical purposes. 

\begin{table}[h!]
  \caption{Ablation study for subsets of components of the SoccerMap architecture.}
  \label{table:ablation}
  \centering
  \begin{tabular}{|p{3.2cm}|p{1.2cm}|p{1.2cm}|p{1.2cm}|p{1.2cm}|p{1.2cm}|p{1.6cm}|}
    \hline
	Architecture & SC & UP & FL & NLP & NF & Log-loss \\
	\hline
	\textbf{SoccerMap}    & YES & YES & YES & YES & 2 & \textbf{0.217} \\
	SoccerMap-NLP & YES & YES & YES & NO & 2 & 0.245 \\
	SoccerMap-FL & YES & YES & NO & YES & 2 & 0.221 \\
	SoccerMap-FL-NLP & YES & YES & NO & NO & 2 & 0.292 \\
	\textbf{SoccerMap-UP} & YES & NO & YES & YES & 2 & \textbf{0.216} \\
	SoccerMap-UP-FL & YES & NO & NO & YES & 2 & 0.220 \\
	SoccerMap-UP-NLP & YES & NO & YES & NO & 2 & 0.225 \\
	SoccerMap-FL-NLP & YES & NO & NO & NO & 2 & 0.235 \\
	Single Layer CNN-D4 & NO & YES & YES & YES  & 2 & 0.256 \\
	Single Layer CNN-D8 & NO & YES & YES & YES  & 4 & 0.228 \\		
	\hline  
\end{tabular}
\end{table}

\section{Practical Applications}
In this section, we present a series of novels practical applications that make use of the full probability surface for evaluating potential passing actions and assessing player’s passing and positional skills.

\subsection{Adapting SoccerMap for the Estimation of Pass Selection Likelihood and Pass Value.}\label{sec:pass_select}

One of the main advantages of this architecture is that is can be easily adapted to other challenging problems associated with the estimation of pass-related surfaces, such as the estimation of pass selection and pass value.

\paragraph{Pass selection model} An interesting and unsolved problem in soccer is the estimation of the likelihood of a pass being made towards every other location on the field, rather than to specific player locations. 
We achieve this by directly modifying the sigmoid activation layer of the original architecture by a softmax activation layer, which ensures that the sum of probabilities on the output surface adds up to $1$. For this case, instead of pass success, we use a sparse matrix as a target output and set the destination location of the pass in that matrix to $1$.

\paragraph{Pass value model} While a given pass might have a low probability of success, the expected value of that pass could be higher than a different passing option with higher probability, thus in some cases, the former could be preferable. We can directly adapt SoccerMap to estimate a pass value surface by modifying the target value and the loss function to be used. For this case, we use as an outcome the expected goals value \cite{eggels2016expected} of the last event in possession of any given pass, which can be positive or negative depending on whether the attacking or defending team had the last action in that possession.

Figure \ref{fig:selection_value} presents the surfaces for pass selection and pass value models derived from this architecture. With these surfaces, we can provide direct visual guidance to coaches to understand the value of the positioning of its team, the potential value gains of off-ball actions, and a team's likely passes in any given situation. 

\begin{figure}[h!]
\centering
\includegraphics[width=1\linewidth]{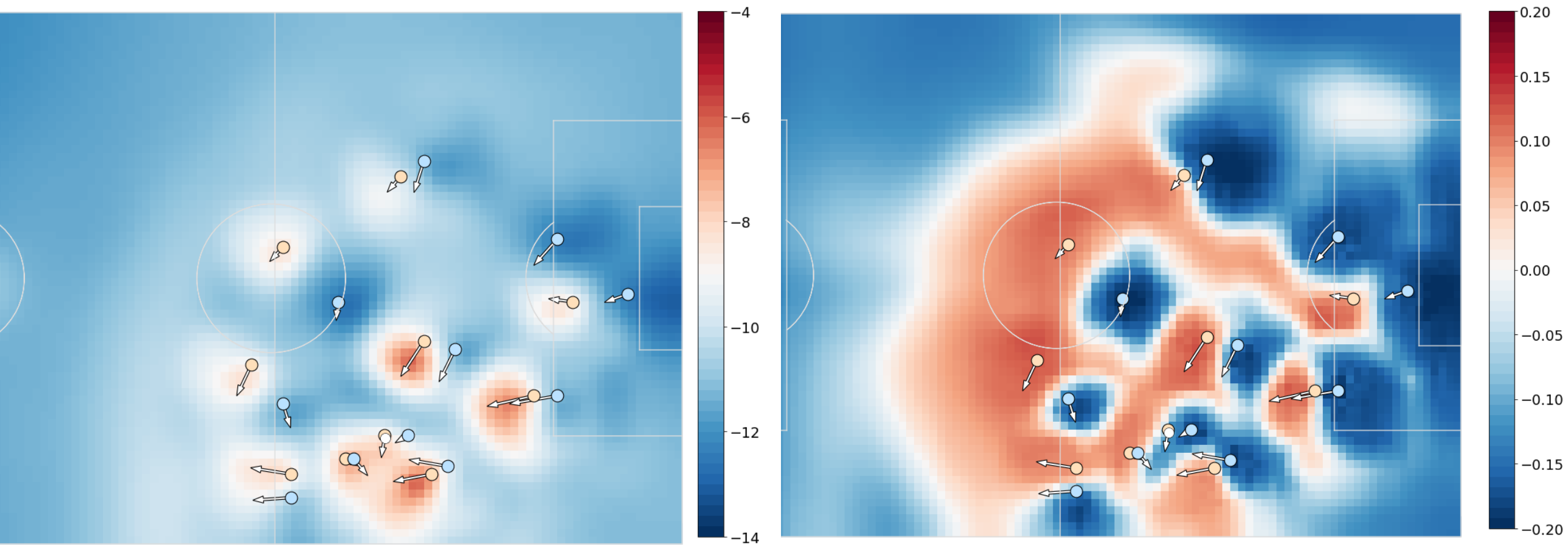}
\caption{On the left column, the pass selection surface for a give game-state, presented on a logarithmic scale. On the right column, a pass value surfaces for the same game-state, with the color scale constrained to a [-0.2,0.2] range.}
\label{fig:selection_value}
\end{figure}

\subsection{Assessing Optimal Passing and Location}

While soccer analytics  have long-focused on using pass probabilities to evaluate a player's passing skills based on observed pass accuracy \cite{power2017not,spearman2017physics}, there are still two main challenging problems that remain unattended: the identification of optimal passing locations and optimal off-ball positioning for improving pass probability.

\subsubsection{Visual Assessment of Optimal Passing}
Given a game-state, where a player is in possession of the ball, we define the optimal and sub-optimal pass destinations as the locations near the teammates than provide a higher pass probability than the current location of the corresponding teammate. To obtain the optimal passing locations we first calculate the pass probability surface of a given game-state and then evaluate the probability of every possible location in a $5 \times 5$ grid around the expected teammate location in the next second, based on the current velocity. The location within that grid with the highest probability difference with the current player's location is set as the optimal passing location. Additionally, a set of sub-optimal passing locations are obtained by identifying locations with positive probability difference and that are at least 5 meters away from previous sub-optimal locations. In the left column of Figure \ref{fig:optimal_location} we present in red circles the set of best passing locations for each of the possession team players for a given game state. This kind of visualization provides a coach the ability to perform a direct visual inspection of passing options and allows her to provide direct feedback to players about specific game situations, improving the coach's effective communication options.

\subsubsection{Visual Assessment of Optimal Positioning}
Following a similar idea, we can leverage pass probabilities surfaces to detect the best possible location a player could occupy to increase the probability of receiving a pass directly. To obtain the optimal location for each player, we recalculate the pass probability surface of the same game situation but translating the location of the player (one player at a time) to any other possible location in the $5 \times 5$ grid. We analogously obtain the optimal locations, as described before. In the right column of Figure \ref{fig:optimal_location} we observe in green circles the expected pass probability added if the player would have been placed in that location instead. Again, this tool can be handy for coaches to instruct players on how to improve their off-ball game.

\begin{figure}[h!]
\centering
\includegraphics[width=1\linewidth]{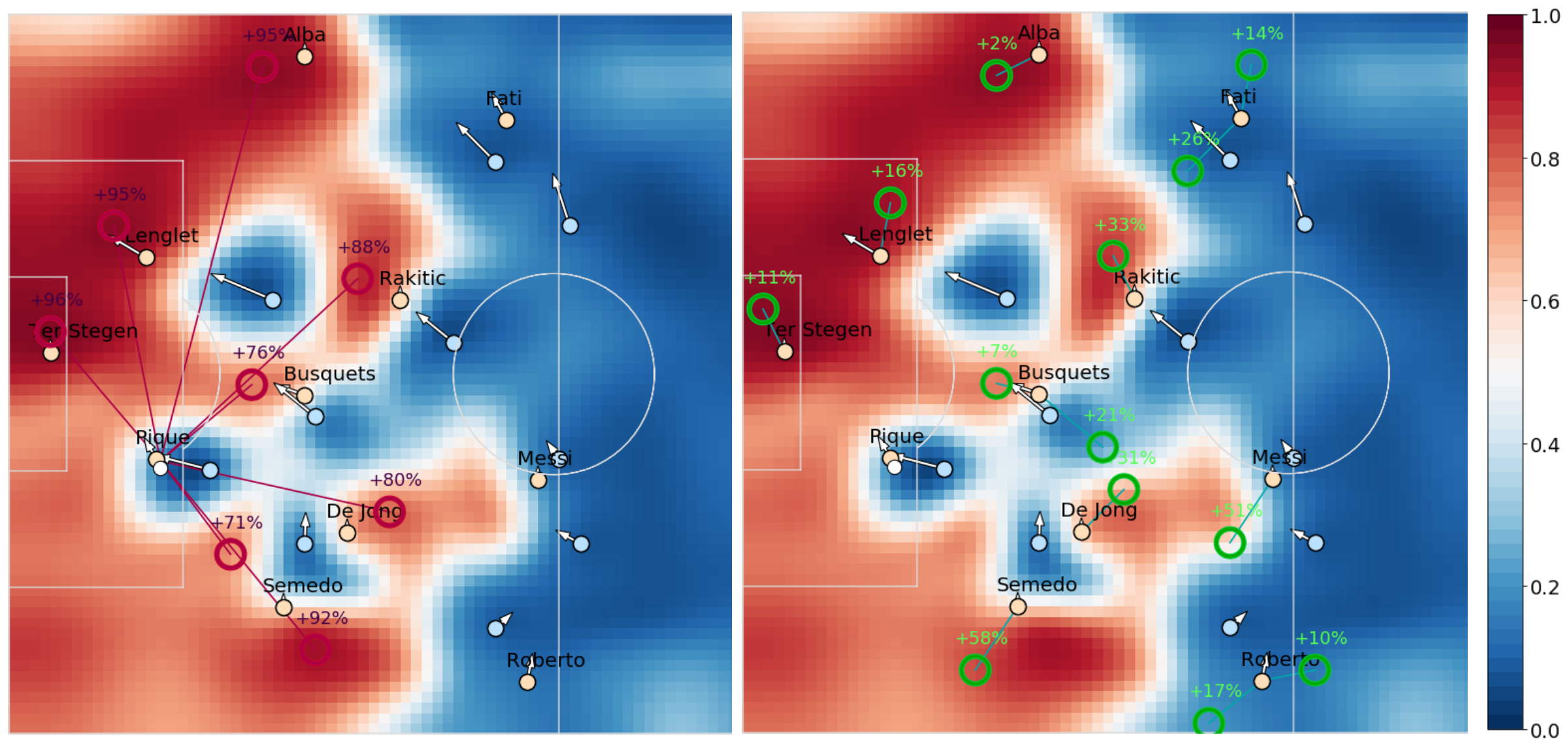}
\caption{In the left column, we present a game-state where red circles represent the optimal passing location for each teammate, and the expected pass probability. In the right column, the green circles represent the optimal positioning of players increasing the expected pass probability if the players were placed in those locations at that time.}
\label{fig:optimal_location}
\end{figure}

\subsubsection{Assessing Passing Skill}
We propose a new metric \textit{pass completion added (PPA)} to evaluate the quality of a players' selection of the passing destination location. For each observed pass, we calculate the difference between the probability of the optimal pass and the probability of the selected pass. This metric is formally defined in Equation \ref{eq:ppa2}, where $S$ and $M$ are the set of successful and missed passes, respectively,  $\hat{y}$ is the optimal pass probability, and $y$ is the selected pass probability. 
Intuitively a player reward is discounted if the selected pass was not optimal. In the case of the pass being unsuccessful, the player is only penalized in proportion to the probability difference with the optimal location, rewarding the player’s pass selection.

\begin{equation}
\label{eq:ppa2}
PPA = \sum_{s=1}^{S}(1-\hat{y}^s)(1-(\hat{y}^s-y^s)) - \sum_{m=1}^M (\hat{y}^s)(\hat{y}^s-y^s)
\end{equation}

In table \ref{table:ppa_table} we present the best ten players in pass completion added for the 2014-2015 season of the English Premier League, where The cumulative $PPA$ of a player is normalized by 90 minutes played. The table includes the estimated player price in 2014, provided by \url{www.transfermarkt.com}. We can observe that the list contains a set of the best players in recent times in this league, including creative midfielders such as Oezil,Silva, Hazard and Fabregas, deep creative wingers such as Navas and Valencia, and Rosicky, a historical player.

\begin{table}[h!]
  \caption{Ranking of the best ten players in pass completion added for the season 2014-2015 of the English Premier League.}
  \label{table:ppa_table}
  \centering
  \begin{tabular}{|c|c|c|c|c|}
    \hline
    Team & Player Name & PPA/90m & Age in 2014 & Player price (2014) \\ \hline
    Arsenal & Mesut Oezil & 0.0578 & 24 & \euro 45M\\
    Manchester City & David Silva & 0.0549 & 28 & \euro 40M\\
    Chelsea & Eden Hazard & 0.0529 & 23 & \euro 48M\\
    Manchester United & Antonio Valencia & 0.0502 & 29 & \euro 13M\\
    Arsenal & Tomas Rosicky & 0.0500 & 33 & \euro 2M\\
    Chelsea & Cesc Fabregas & 0.0484 & 27 & \euro 40M\\
    Arsenal & Santi Cazorla & 0.0470 & 29 & \euro 30M\\
    Manchester City & Jesus Navas & 0.0469 & 28 & \euro 20M\\
    Manchester City & Yaya Toure & 0.0466 & 30 & \euro 30M\\
    Manchester City & Samir Nasri & 0.0447 & 26 & \euro 22M \\
	\hline  
\end{tabular}
\end{table}

\subsection{Team-Based Passing Selection Tendencies}

The pass selection adaptation of SoccerMap, presented in Section \ref{sec:pass_select}, provides a fine-grained evaluation of the passing likelihood in different situations. However, it is clear to observe that passing selection is likely to vary according to a team's player style and the specific game situation. While a league-wide model might be useful for grasping the expected behavior of a typical team in the league, a soccer coach will be more interested in understanding the fine-grained details that separate one team from the other. Once we train a SoccerMap network to obtain this league-wide model, we can fine-tune the network with passes from each team to grasp team-specific behavior. In this application example, we trained the pass selection model with passes from all the teams from English Premier League season 2014-2015. Afterward, we retrained the initial model with passes from two teams with different playing-styles: Liverpool and Burnley. \\

In Figure \ref{fig:team_pass_selection} we compare the pass selection tendencies between Liverpool (left column) and Burnley (right column). On the top left corner of both columns, we show a 2D plot with the difference between the league mean passing selection heatmap, and each team's mean passing selection heatmap, when the ball is within the green circle area. We can observe that Liverpool tends to play short passes, while Burnley has a higher tendency of playing long balls to the forwards or opening on the sides.  However, this kind of information would not escape from the soccer coach's intuition, so we require a more fine-grained analysis of each team's tendencies in specific situations. In the two plots of Figure \ref{fig:team_pass_selection} we show over each players' location the percentage increase in passing likelihood compared with the league's mean value. In this situation, we can observe that when a left central defender has the ball during a buildup, Liverpool will tend to play short passes to the closest open player, while Burnley has a considerably higher tendency to play long balls to the forwards, especially if forwards are starting a run behind the defender's backs, such as in this case. Through a straightforward fine-tuning of the SoccerMap-based model, we can provide detailed information to the coach for analyzing specific game situations.

\begin{figure}[h!]
\centering
\includegraphics[width=1\linewidth]{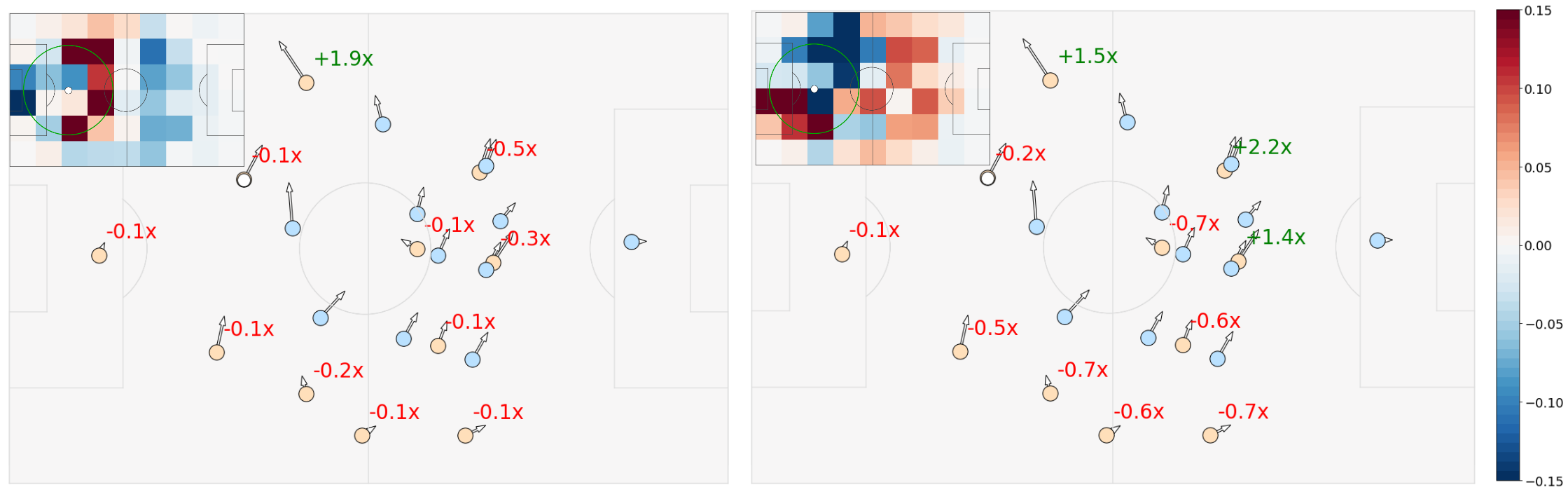}
\caption{A game-state representation of a real game situation in soccer. Above each player (circles) we present the added percentage difference of pass likelihood in that given situation in comparison with the league for two teams: Liverpool (left column) and Burnley (right column). The heatmaps in both top left corners of each column represent the mean difference in pass selection likelihood with the league, when the ball is located within the green circle.}
\label{fig:team_pass_selection}
\end{figure}

\section{Discussion and Future Work}

The estimation of full probability surfaces provides a new dimension for soccer analytics.  The presented architecture allows generating visual tools to help
coaches perform fine-tuned analysis of opponents and own-team performance, derived from low-level spatiotemporal soccer data. We show how this network can be easily adapted to many other challenging related problems in soccer, such as the estimation of pass selection likelihood and pass value, and that can perform remarkably well at estimating the probability of observed passes. By merging features extracted at different sampling levels, the network can extract both fine and coarse details, thereby managing to make sense of the complex spatial dynamics of soccer. We have also presented several novels practical applications on soccer analytics, such as evaluating optimal passing,
evaluating optimal positioning, and identifying context-specific and team-level passing tendencies. This framework of analysis derived from spatiotemporal data could also be applied directly in many other team sports, where the visual representation of complex information can bring the coach and the data analyst closer.


\end{document}